\documentclass[10pt,twocolumn,letterpaper]{article}

\usepackage[pagenumbers]{cvpr} 

\definecolor{cvprblue}{rgb}{0.21,0.49,0.74}
\usepackage[pagebackref,breaklinks,colorlinks,allcolors=cvprblue]{hyperref}

\definecolor{codegreen}{rgb}{0,0.6,0}
\definecolor{codegray}{rgb}{0.5,0.5,0.5}
\definecolor{codepurple}{rgb}{0.58,0,0.82}
\definecolor{backcolour}{rgb}{0.95,0.95,0.92}
\definecolor{LightBlue}{HTML}{f2f2f2}
\definecolor{baseCustomColor}{HTML}{9B7EBE}
\colorlet{stepColorOne}{baseCustomColor}
\colorlet{stepColorTwo}{baseCustomColor!90!blue}
\colorlet{stepColorThree}{baseCustomColor!80!blue}
\colorlet{stepColorFour}{baseCustomColor!60!blue}

\newcommand{\bolditn}[1]{\textbf{\textsl{#1}}}

\def\paperID{} 
\def\confName{CVPR}
\def\confYear{2026}

\usepackage[utf8]{inputenc} 
\usepackage[T1]{fontenc}    
\usepackage[dvipsnames,table]{xcolor}
\usepackage{hyperref}       
\usepackage{url}            
\usepackage{booktabs}       
\usepackage{amsfonts}       
\usepackage{nicefrac}       
\usepackage{microtype}      
\usepackage{enumitem}
\usepackage{pdfcomment}

\usepackage{bm}
\usepackage{amsmath} 
\usepackage{amssymb} 
\usepackage{graphicx} 
\usepackage{xcolor}
\usepackage{multirow} 
\usepackage{subcaption} 
\usepackage{colortbl}  
\usepackage{wrapfig}
\usepackage{pifont}

\title{
Reason-SVG: Enhancing Structured Reasoning for Vector Graphics Generation with Reinforcement Learning
}

\author{%
  Ximing Xing$^{1}$,
  Ziteng Xue$^{1}$,
  Yandong Guan$^{1}$,
  Jing Zhang$^{1}$,
  Dong Xu$^{2}$,
  Qian Yu$^{1}$\thanks{Corresponding author} \\
  $^{1}$Beihang University \quad $^{2}$The University of Hong Kong \\
  \texttt{\{ximingxing, qianyu\}@buaa.edu.cn} \quad
  \texttt{dongxu@cs.hku.hk}
}

\begin{document}

\maketitle

\begin{abstract}
Generating high-quality Scalable Vector Graphics (SVGs) is challenging for Large Language Models (LLMs), as it requires advanced reasoning for structural validity, semantic accuracy, and visual coherence—areas where current LLMs often struggle. In this work, we introduce Reason-SVG, a novel framework equipped with enhanced structured reasoning for SVG generation. Reason-SVG pioneers the ``Drawing-with-Thought'' (DwT) paradigm, in which models generate both SVG code and explicit design rationales.
Reason-SVG follows a two-stage training strategy: First, Supervised Fine-Tuning (SFT) trains the LLM on the DwT paradigm to develop foundational reasoning abilities. Second, Reinforcement Learning (RL), utilizing Group Relative Policy Optimization (GRPO), empowers the model to generate both DwT and SVG rationales through refined, reward-driven reasoning.
To enable reasoning-driven SVG generation, we design a Hybrid Reward function that evaluates the presence and effectiveness of DwT reasoning, along with structural validity, semantic alignment, and visual quality. We also introduce the SVGX-DwT-10k dataset, a high-quality corpus of 10k SVG-DwT pairs, where each SVG code is generated based on explicit DwT reasoning.
By integrating DwT, SFT, and Hybrid Reward-guided RL, Reason-SVG significantly improves the performance of LLMs and VLMs in generating accurate and visually coherent SVGs.
\end{abstract}

\section{Introduction}
Scalable Vector Graphics (SVG) offer lossless scalability and editability, advantages that have led to their widespread adoption in applications from font design~\cite{svgvae_lopes_2019,deepvecfont_wang_2021,vecfusion_thamizharasan_2024} to data visualization~\cite{svgdatavis_xu_2024,chart4blind_moured_2024}. As an XML-based language, an SVG has a dual nature: it is simultaneously a visual graphic and structured source code. In recent years, Text-to-SVG generation has garnered significant attention. However, the task is challenging because the output must satisfy both visual and code criteria: being aesthetically pleasing while also well-structured and editable.

\begin{figure*}[t]
\centering
\includegraphics[width=1.0\textwidth]{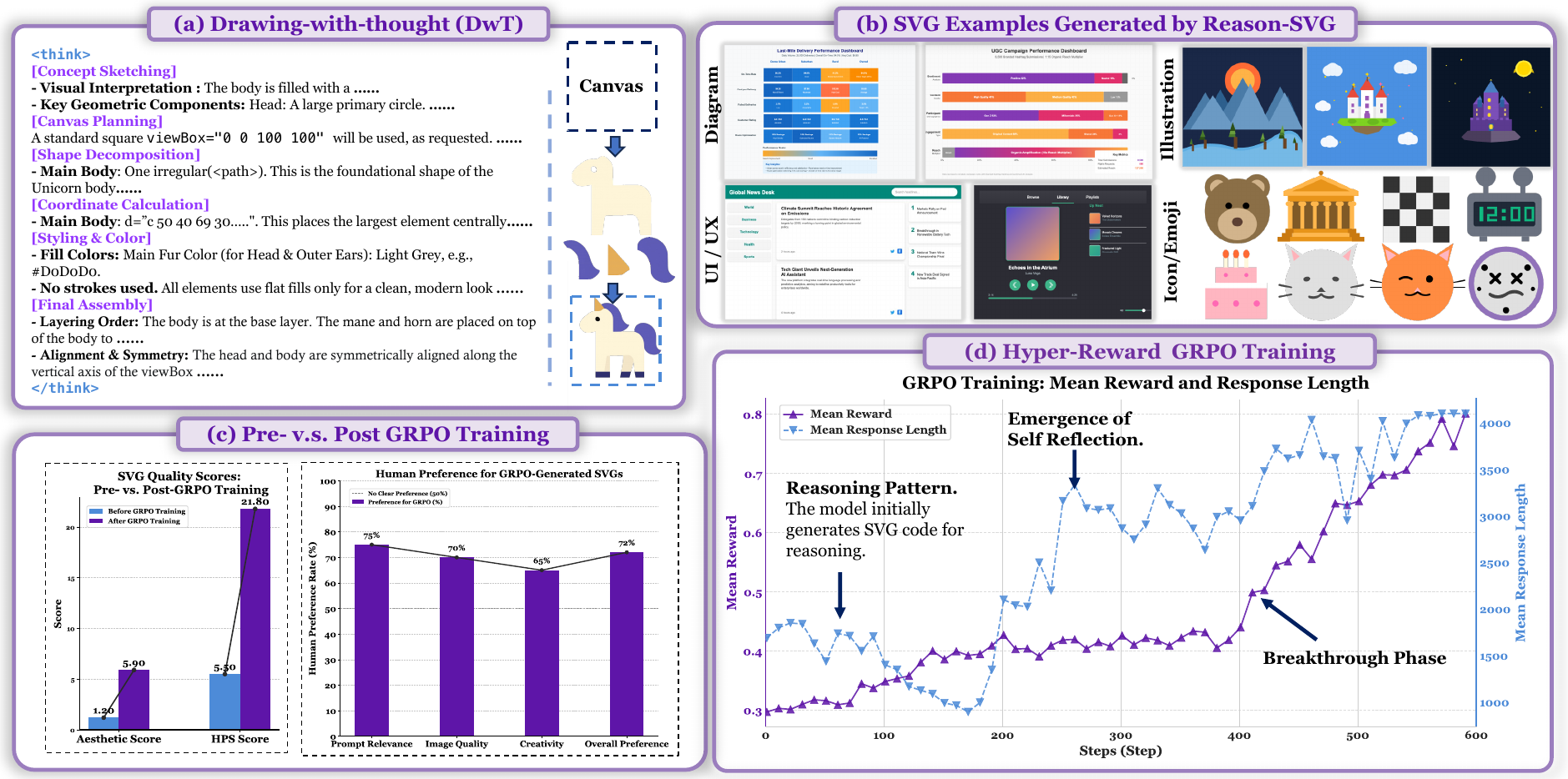}
\vspace{-1em}
\caption{
\textbf{Overview of} \bolditn{Reason-SVG}. 
Reason-SVG incorporates structured reasoning through the \bolditn{Drawing-with-Thought} (DwT) paradigm, enabling LLMs to synthesize SVGs guided by explicit visual planning and compositional logic.  
\textbf{\color{Plum}{(a)} DwT Reasoning Process:}  
An example of the Drawing-with-Thought reasoning process, illustrating structured design decisions across stages such as conceptual design, preliminary design, and detailed design.
\textbf{\color{Plum}{(b)} SVG Samples:}  SVGs generated by Reason-SVG demonstrate superior compositional quality and accurate spatial layout, confirming enhanced capability for complex prompts and visually coherent graphics.
\textbf{\color{Plum}{(c)} Quantitative Improvements:}  
GRPO training significantly enhances visual aesthetics, semantic fidelity, and human preference scores across multiple evaluation dimensions. 
\textbf{\color{Plum}{(d)} Optimization Insight – GRPO Training:}  
During GRPO training, the model gradually learns that longer and more structured responses tend to receive higher rewards, revealing an implicit coupling between reward signals and response length.   
} \label{fig:teaser}
\vspace{-1em}
\end{figure*}

Existing SVG generation methods fall into two main paradigms. The first, optimization-based approaches~\cite{diffvg_Li_2020,clipdraw_frans_2022,vectorfusion_jain_2023,diffsketcher_xing_2023,vectorpainter_hu_2024,svgdreamer_xing_2023,neualpath_zhang_2024,nivel_thamizharasan_2024}, iteratively refine SVG parameters under the guidance of CLIP~\cite{clip_Radford_2021} or T2I models to achieve high visual fidelity. However, this process is computationally intensive and often yields poorly editable SVG code.
In contrast, the second paradigm leverages Large Language Models (LLMs)~\cite{iconshop_wu_2023,strokenuwa_tang_2024,starvector_Rodriguez_2023,llm4svg_xing_2024,starcoder_li_2023,rendering_rodriguez_2025,unisvg_li_2025,svggen_wang_2025,omnisvg_yang_2025,internsvg_wang_2025} by treating SVG creation as a code generation task. This approach significantly improves generation speed while producing more structured, editable code, establishing it as a promising direction. Despite their potential, current LLM-based methods are often limited by a poor understanding of complex semantics and a tendency to overfit training data.
For instance, while existing methods can readily generate a high-quality SVG for a simple prompt like ``a castle'', they typically fail to produce coherent or accurate results for a more complex compositional prompt such as ``a white-and-red castle on a floating island among clouds in a blue sky'' (the 2nd example shown in Fig.~\ref{fig:teaser}(d) \textit{Illustration}).


We posit that this core limitation arises from the inherent ambiguity and high complexity of mapping a concise, high-level textual prompt directly to a verbose, low-level SVG code. 
While LLMs are pre-trained on vast repositories of web data containing SVG/XML snippets, this training lacks the explicit, fine-grained annotations that link semantic concepts within a prompt (e.g., `castle', `island', `clouds', `sky', and their relationships) to specific structural elements and attributes in the code. This creates a significant semantic gap, forcing the model to learn this complex mapping implicitly. 


To bridge this semantic gap, we propose an intermediate reasoning process that acts as a conceptual scaffold between the prompt and the final code. Our approach is motivated by the idea of creating a concrete plan before generation. Instead of attempting a direct translation, we first decompose the complex prompt into a structured, high-level plan that explicitly outlines key semantic components, their hierarchical and spatial relationships, and their stylistic attributes. Generating this intermediate plan effectively simplifies the overall task into two more manageable sub-problems: first, reasoning about what to draw and how to arrange it conceptually, and second, translating that well-defined plan into valid SVG code.

Building on the analysis above, we introduce \textbf{\textit{Reason-SVG}}, a novel framework equipped with a reasoning process named "\textbf{Drawing-with-Thought}" (\textbf{DwT}). Given an input text prompt, the model first generates a detailed DwT rationale.
This rationale serves as a blueprint, explicitly decomposing the prompt into its core conceptual components (\textit{Conceptual Design}), outlining their spatial arrangement and structural roles (Preliminary Design), and planning for final attributes like color and style (Detailed Design). 
By conditioning the final code generation on this explicit and structured thought process, we transform an ill-posed, high-level generation task into a more tractable, step-by-step rendering process. This enables the model to robustly handle intricate semantic relationships that it would otherwise fail to capture.



While this structured DwT provides a strong inductive bias for reasoning, a single, predefined reasoning template may not be optimal for all cases. To allow the model to discover and refine its own reasoning pathways, we introduce a two-stage training strategy. First, we perform Supervised Fine-Tuning (SFT) on an LLM using a curated dataset of DwT-annotated SVGs. This stage teaches the model to generate SVG code concurrently with an explicit reasoning trace. Second, building on this foundation, we employ Generative Retraining with Policy Optimization (GRPO)~\cite{deepseekr1_guo_2025}, a reinforcement learning (RL) technique, to further refine the model. This RL stage encourages the model to explore the generation space, optimizing for both more effective DwT rationales and higher-quality final SVGs.

A key challenge in applying RL to this task is the lack of a single ``correct'' output; both the final SVG and the underlying reasoning can be valid in many forms. Simple, rule-based rewards are insufficient to capture this complexity. To address this, we design and implement a novel \textbf{Hybrid Reward function}. This function provides comprehensive feedback by jointly evaluating four critical aspects: (1) the structural validity of the generated SVG code, (2) the semantic alignment between the SVG and the input prompt, (3) the aesthetic quality of the rendered image, and, crucially, (4) the logical coherence of the DwT rationale itself.

To support our research, we construct and release SVGX-DwT-10k, a large-scale dataset comprising 10,000 high-quality SVGs paired with DwT rationales that are verified and refined by an LLM guided by a carefully designed system prompt. Our contributions are threefold:
\begin{itemize}
\item We propose Reason-SVG, a novel framework that introduces a Drawing-with-Thought (DwT) process to instill explicit reasoning in LLM-based SVG generation. 
\item We design a two-stage training pipeline combining SFT for initial reasoning alignment and RL-based refinement (GRPO) guided by a novel Hybrid Reward function that evaluates both the final output and the reasoning process.
\item We introduce the SVGX-DwT-10k dataset to facilitate research into reasoning-driven SVG generation. 
We conduct extensive experiments to demonstrate that the proposed DwT and Hybrid Reward are also applicable in VLM-based SVG generation.
\end{itemize}

\section{Related Work}

\subsection{Vector Graphics Generation}
Research on SVGs spans generation and understanding of vector structures. Early neural approaches model SVG command sequences with RNNs/Transformers/VAEs and, more recently, diffusion~\cite{sketchrnn_david_2018,svgvae_lopes_2019,im2vec_reddy_2021,deepsvg_carlier_2020,deepvecfont_wang_2021,iconshop_wu_2023,strokenuwa_tang_2024,beyondpixels_zhang_2023,supersvg_hu_2024,xing2024svgfusion}, but progress is limited by the scarcity of diverse, well-annotated vector corpora. A complementary line adopts differentiable rasterization~\cite{diffvg_Li_2020} to optimize SVG parameters with CLIP- or diffusion-guided objectives~\cite{im2vec_reddy_2021,live_Ma_2022,supersvg_hu_2024,clip_Radford_2021,clipdraw_frans_2022,clipasso_vinker_2022,clipascene_vinker_2023,clipvg_song_2023,clipgen_shen_2022,vectorfusion_jain_2023,diffsketcher_xing_2023,wordasimg_iluz_2023,svgdreamer_xing_2023,svgdreamerplus_xing_2025}. Recent works further explore neural shape priors~\cite{nivel_thamizharasan_2024,neualpath_zhang_2024,neuralsvg_polaczek_2025}, personalization~\cite{svgcustomization_zhang_2023}, and stylization~\cite{vectorpainter_hu_2024}. On the understanding side, vector-native recognizers~\cite{yolat_jiang_2021,yolat++_dou_2024} and LLM-oriented benchmarks~\cite{svgeditbench_nishina_2024,vgbench_zou_2024} reveal that, despite good code-level comprehension, generation and editing often degrade on complex geometry.


\subsection{Drawing with Large Language Models}
LLMs exhibit strong language understanding and generalization~\cite{gpt4_report,claude3.5,claude3.7,qwen2.5_yang_2024,deepseekv3_liu_2024,deepseekr1_guo_2025,o4mini}. Benchmarks assess their SVG parsing/editing abilities~\cite{vgbench_zou_2024,svgeditbench_nishina_2024,pvd_wang_2024}, while systems like Chat2SVG~\cite{chat2svg_wu_2024} use LLMs to propose semantic templates for optimization pipelines. To strengthen SVG synthesis, many works curate data and apply SFT~\cite{iconshop_wu_2023,strokenuwa_tang_2024,starvector_Rodriguez_2023,llm4svg_xing_2024,omnisvg_yang_2025}, including tokenization and structure–geometry decoupling; concurrent efforts expand training sets and code-style generation~\cite{unisvg_li_2025,svggen_wang_2025}. By constructing specialized SVG datasets and fine-tuning mainstream LMs and Coder-LMs, the ability to generate synthetic SVGs is being actively pursued. Reinforcement Learning with rendering feedback has also been explored to refine SVG outputs~\cite{rendering_rodriguez_2025}. In contrast, we employ a GRPO-based reasoning objective that explicitly trains DwT-style planning before code emission, yielding more coherent, compositional SVGs.

\section{Preliminary}
\label{sec:preliminary}

\subsection{SFT-based SVG Generation}
\label{sec:preliminary_sft}

Supervised Fine-Tuning (SFT)~\cite{training_ouyang_2022,selfinstruct_wang_2022,llava_liu_2023} is a standard technique for adapting Large Language Models (LLMs) to downstream tasks such as SVG synthesis.  
This process involves training LLMs on specialized datasets consisting of (input, SVG code) pairs to instill domain-specific knowledge.  
Several recent works~\cite{starvector_Rodriguez_2023,llm4svg_xing_2024,omnisvg_yang_2025} leverage SFT to improve the SVG generation capabilities of LLMs.

The SFT objective typically maximizes the likelihood of a target SVG token sequence $y = (y_1, \dots, y_T)$ given an input condition $\bm{x}_{\text{cond}}$, which may consist of an instruction $\bm{x}_{\text{inst}}$~\cite{llm4svg_xing_2024}, and optionally include other modalities such as an image $\bm{x}_{\text{img}}$~\cite{starvector_Rodriguez_2023} or an SVG embedding $\bm{x}_{\text{svg}}$~\cite{omnisvg_yang_2025}:
\vspace{-0.5em}
\begin{equation}
\mathcal{L}_{\text{SFT}} = -\mathbb{E}_{(\bm{x}_{\text{cond}}, y) \sim \mathcal{D}} \sum_{t=1}^{T} \log \pi_{\theta}(y_t \mid \bm{x}_{\text{cond}}, y_{<t}),
\label{eq:sft_loss}
\vspace{-0.5em}
\end{equation}
where $\mathcal{D}$ denotes the fine-tuning dataset (e.g., SVG-Stack~\cite{starvector_Rodriguez_2023}, SVGX-SFT~\cite{llm4svg_xing_2024}, MMSVG-Icon~\cite{omnisvg_yang_2025}), and $\pi_{\theta}$ is the LLM parameterized by $\theta$.  
The conditioning input $\bm{x}_{\text{cond}}$ varies across methods, including visual features~\cite{starvector_Rodriguez_2023} and specialized token representations~\cite{llm4svg_xing_2024,omnisvg_yang_2025}.

\subsection{Group Relative Policy Optimization}
\label{sec:preliminary_rl}
Reinforcement Learning (RL) is widely used to enhance LLM reasoning for structured, multi-step generation~\cite{deepseekr1_guo_2025,o4mini,unifiedreward_wang_2025,r1reward_zhang_2025,reasonrft_tan_2025,dwt_cui_2025}. A common choice is Proximal Policy Optimization (PPO)~\cite{ppo_schulman_2017}, which stabilizes updates via a clipped surrogate objective.

Group Relative Policy Optimization (GRPO)~\cite{deepseekr1_guo_2025}, popularized by DeepSeek-R1, adapts PPO to rule-based rewards where explicit supervision is unavailable. It estimates advantages by comparing multiple trajectories sampled from the current policy instead of using a learned value function, making it well-suited to deterministic or heuristic reward signals.
GRPO maximizes a clipped objective with a KL penalty toward a reference policy:
$$
J_{\text{GRPO}}(\theta)=\mathbb{E}_{q,\{a_i\}}\!\left[\left\langle L_{\text{clip}}(\theta,i,t)\right\rangle-\beta\,D_{\text{KL}}\!\left(\pi_\theta \,\|\, \pi_{\text{ref}}\right)\right],
$$
where $\langle\cdot\rangle$ averages over $G$ responses and their tokens, $q\!\sim\!\mathcal{D}$, and $\{a_i\}_{i=1}^G\!\sim\!\pi_{\theta_{\text{old}}}$. The clipped loss is
$
L_{\text{clip}}(\theta,i,t)=\min\!\big(r_{i,t}(\theta)\hat{A}_{i,t},\,\text{clip}(r_{i,t}(\theta),1-\epsilon,1+\epsilon)\hat{A}_{i,t}\big),
$
with probability ratio
\vspace{-0.5em}
\begin{equation*}
\small
r_{i,t}(\theta)=\frac{\pi_\theta(o_{i,t}\mid q,o_{i,<t})}{\pi_{\theta_{\text{old}}}(o_{i,t}\mid q,o_{i,<t})}.
\end{equation*}
This formulation enables policy optimization from rule-based or heuristic rewards without dense token-level supervision.

Nevertheless, open-ended Text-to-SVG is complex and under-specified: there is no single ground truth, and purely rule-based rewards often lack fine-grained guidance across planning and rendering, limiting RL effectiveness without additional structure.

\begin{figure*}[h]
\centering
\includegraphics[width=\textwidth]{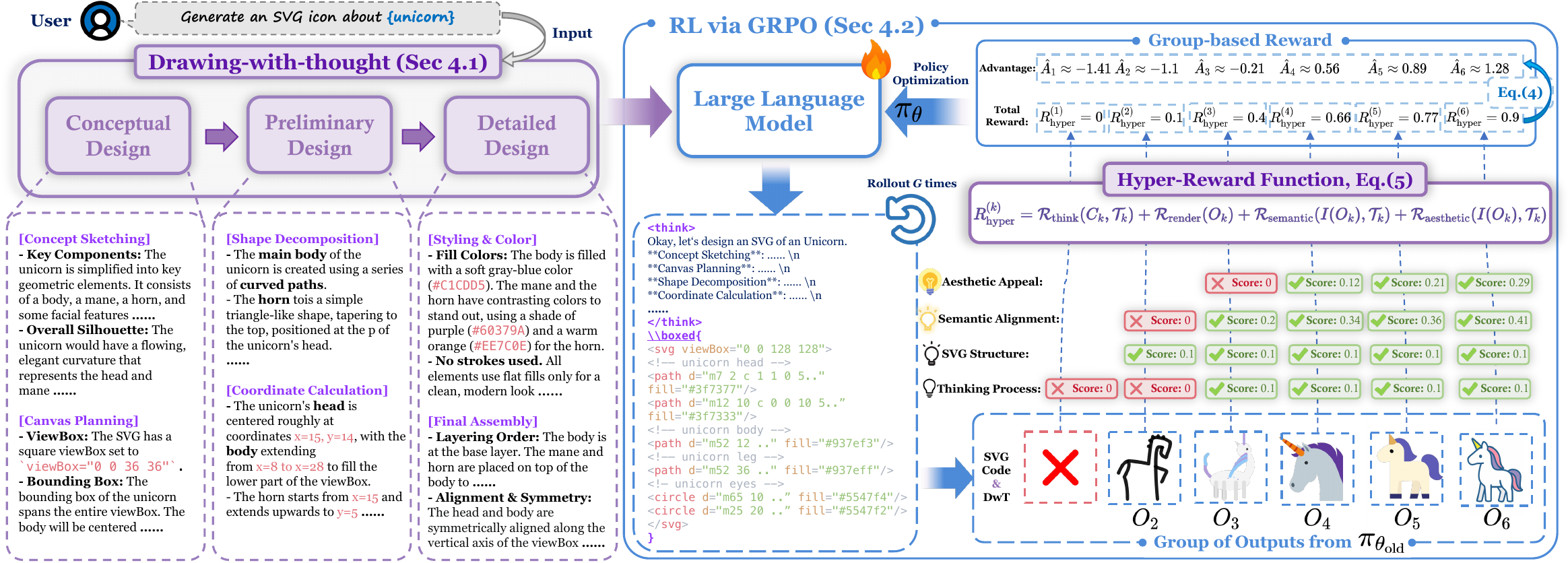}
\vspace{-1em}
\caption{
\textbf{Framework of}~\bolditn{Reason-SVG}.  
The ``Drawing-with-Thought'' (DwT, Sec.~\ref{sec:dwt}) module guides the LLM through a step-by-step visual reasoning process to generate both the SVG code ($O$) and its corresponding design rationale ($C$).  
This process comprises the following stages:  
\textit{a)} concept sketching,  
\textit{b)} canvas planning,  
\textit{c)} shape decomposition,  
\textit{d)} coordinate calculation,  
\textit{e)} styling and coloring, and  
\textit{f)} final assembly.  
These reasoning stages culminate in a coherent SVG output, which is subsequently refined via reinforcement fine-tuning (RFT) using a Hybrid Reward (Sec.~\ref{sec:hybrid_reward}) that jointly evaluates semantic alignment, visual aesthetics, and structural validity.
}
\label{fig:method_overview}
\vspace{-1em}
\end{figure*}
\section{Reason-SVG}
\label{sec:method}
In this section, we present \bolditn{Reason-SVG}, a novel framework designed to enhance the reasoning capabilities of LLMs in vector graphics generation.  
Reason-SVG adopts a ``planning-then-drawing'' paradigm, where the model is first guided to produce an explicit design rationale—a ``Drawing-with-Thought'' (DwT) sequence—followed by the generation of the corresponding SVG code.  
This is implemented through a two-stage training pipeline:  
(1) Supervised Fine-Tuning (SFT) to activate the model’s reasoning ability via DwT supervision, and  
(2) Reinforcement Learning (RL) with a novel hybrid reward function to jointly refine both the reasoning process and the final SVG output.  
Figure~\ref{fig:method_overview} provides an overview of the Reason-SVG architecture.

\subsection{Drawing-with-Thought (DwT)}
\label{sec:dwt}
The core idea of Drawing-with-Thought (DwT) is to enable an LLM to explicitly generate a chain of reasoning steps, denoted as $C$ (the ``thought''), prior to producing the final SVG code $O$, based on an input textual description $\mathcal{T}$.  
This process mimics how human designers typically conceptualize and plan before executing a design.  
The overall generation procedure can be formalized as a mapping $\Phi: \mathcal{T} \rightarrow (C, O)$.

\noindent\textbf{The DwT Reasoning Process.}  
The DwT mechanism instantiates a structured reasoning process that emulates the typical workflow of human designers.  
As illustrated in Fig.~\ref{fig:method_overview} (left), this process decomposes the generation of SVG graphics into six sequential stages:  
\textit{(a) Concept Sketching}, which identifies salient visual components (e.g., body, mane, horn) and outlines the overall silhouette;  
\textit{(b) Canvas Planning}, which establishes a standardized \texttt{viewBox} (e.g., \texttt{0 0 100 100}) and defines the spatial layout;  
\textit{(c) Shape Decomposition}, which breaks the composition into geometric primitives (e.g., circles, curves);  
\textit{(d) Coordinate Calculation}, which determines approximate spatial positions for each component;  
\textit{(e) Styling and Coloring}, which assigns a flat color palette and consistent stylistic properties;  
and \textit{(f) Final Assembly}, which integrates all elements into a coherent and visually aligned design.  
This hierarchical reasoning formulation enhances both the interpretability and controllability of SVG generation, and provides a structured foundation for downstream reward-based optimization.

\noindent\textbf{Drawing-with-Thought Reasoning Activation.}  
To equip the model with structured visual reasoning capabilities, we introduce the \bolditn{Drawing-with-Thought} (DwT) mechanism during the supervised fine-tuning (SFT) phase.  
Specifically, we construct a training dataset $\mathcal{D}_{\text{SFT-DwT}}$, where each instance consists of a textual prompt $\mathcal{T}_j$, an expert-authored DwT reasoning sequence $C_j$—structured across six predefined stages—and a corresponding ground-truth SVG output $O_j$.  

During SFT, the language model $\pi_\theta$ takes $\mathcal{T}_j$ as input and is trained to generate a concatenated target sequence comprising the DwT reasoning steps $C_j$ followed by the SVG code $O_j$, in an autoregressive manner.
The training objective is to maximize the conditional likelihood of the complete sequence, thereby encouraging the model to first articulate a coherent, high-level design rationale and then translate it into a well-structured SVG representation.  
This training strategy activates the model’s latent reasoning ability and aligns its generation process with the step-wise decomposition characteristic of design workflows.

\subsection{Hybrid Reward Function Design}
\label{sec:hybrid_reward}
Following SFT, we apply reinforcement learning to further improve the LLM’s ability to generate high-quality Drawing-with-Thought sequences and corresponding SVG code.  
To this end, we utilize Group Relative Policy Optimization (GRPO)~\cite{deepseekr1_guo_2025}, a variant of Proximal Policy Optimization (PPO)~\cite{ppo_schulman_2017} that estimates advantages in a group-relative manner without relying on an explicit value function.

Given a prompt $\mathcal{T}$, the current policy $\pi_{\theta}$ (initialized from the SFT-trained model) generates a set of $G$ diverse candidate sequences $\{A_k = (C_k, O_k)\}_{k=1}^G$, where each $A_k$ consists of a DwT reasoning trace $C_k$ and its corresponding SVG output $O_k$.

The advantage $\hat{A}_{k}$ for each candidate $A_k$ is computed by comparing its total hybrid reward $R_{\text{hybrid}}^{(k)}$ against the overall group performance:
\begin{equation}
\hat{A}_{k} = \frac{R_{\text{hybrid}}^{(k)} - \text{mean}(\{R_{\text{hybrid}}^{(j)}\}_{j=1}^G)}{\text{std}(\{R_{\text{hybrid}}^{(j)}\}_{j=1}^G) + \delta},
\label{eq:grpo_advantage_method}
\end{equation}
where $\delta$ is a small constant for numerical stability.  
The computed advantage is uniformly applied to all tokens in $A_k$ and used to update the policy via the GRPO objective.

To effectively guide Reason-SVG, we design a novel hybrid reward function $\mathcal{R}_{\text{hyper}}$. For each generated candidate $(C_k, O_k)$ from prompt $\mathcal{T}_k$, the total reward $R_{\text{hyper}}^{(k)}$ is defined as a weighted sum of four components:
\begin{equation}
\begin{split}
R_{\text{hyper}}^{(k)} = \lambda_t {\color{stepColorOne}\underbrace{\color{black}\mathcal{R}_{\text{think}}(C_k, \mathcal{T}_k)}_{\text{Thought Process}}} + \lambda_r {\color{stepColorTwo}\underbrace{\color{black}\mathcal{R}_{\text{render}}(O_k)}_{\text{Structural Validity}}} \\
+ \lambda_s {\color{stepColorThree}\underbrace{\color{black}\mathcal{R}_{\text{semantic}}(I(O_k), \mathcal{T}_k)}_{\text{Semantic Alignment}}} + \lambda_a {\color{stepColorFour}\underbrace{\color{black}\mathcal{R}_{\text{aesthetic}}(I(O_k), \mathcal{T}_k)}_{\text{Visual Aesthetic}}}
\label{eq:hyper_reward}
\end{split}
\end{equation}
where $I(O_k)$ denotes the rasterized image rendered from SVG $O_k$, and the non-negative coefficients $\lambda_t, \lambda_r, \lambda_s, \lambda_a$ control the relative importance of each reward term.
The individual reward components are defined as follows:

\noindent\textbf{Thought Process Reward ($\mathcal{R}_{\text{think}}$):}  
This component evaluates whether the generated DwT sequence $C_k$ adheres to the required multi-stage structure by detecting the presence of expected \texttt{<think>} tags.  
Instead of directly assessing the content of each reasoning step, we adopt a lightweight structure-based proxy that only enforces the correct use of structural markers.

\noindent\textbf{SVG Structural Validity Reward ($\mathcal{R}_{\text{render}}$):}  
This component checks whether the generated SVG code $O_k$ is syntactically valid by verifying its renderability using CairoSVG~\cite{cairosvg}.  
The reward is implemented as a binary signal: it returns $1$ if the SVG can be successfully rendered without any syntax or parsing errors, and $0$ otherwise.  
This ensures that the generated outputs comply with SVG grammar and remain functional in downstream usage scenarios.

\noindent\textbf{Semantic Alignment Reward ($\mathcal{R}_{\text{semantic}}$):}\quad
This term evaluates concept-level agreement between the rendered image $I(O_k)$ and the prompt $\mathcal{T}_k$ using CLIP~\cite{clip_Radford_2021}. We obtain normalized image and text embeddings and compute their cosine similarity as a scalar reward. Higher scores indicate stronger semantic consistency—capturing object identity, attributes, and relations—so the SVG reflects the user’s intent beyond mere visual plausibility.

\noindent\textbf{Visual Aesthetic Reward ($\mathcal{R}_{\text{aesthetic}}$):}  
This component directly encourages the generation of visually attractive and professionally styled outputs.  
We employ the HPSv2~\cite{HPS_Wu_2023}, which predicts human-perceived aesthetic preferences based on image–prompt pairs.  
This reward guides the model toward outputs with superior color harmony, visual balance, and overall compositional quality, enhancing the appeal and usability of the generated graphics.

Overall, the hybrid reward function provides rich, multi-dimensional supervision that balances structural correctness, semantic relevance, visual quality, and reasoning completeness.  
Notably, by leveraging structured tags as a proxy for intermediate reasoning, the framework introduces an effective yet simple mechanism to guide the model toward interpretable and purposeful generation.

\begin{table*}[t]
\centering
\resizebox{\textwidth}{!}{%
\begin{tabular}{lccccccccc}
\toprule
Method & Time(s) $\downarrow$ & \#Token & \#Complex & Val\% $\uparrow$ & FID $\downarrow$ & CLIPScore $\uparrow$ & Aesthetic $\uparrow$ & HPSv2 $\uparrow$ & DwT-Cover\% $\uparrow$ \\
\midrule
\multicolumn{10}{l}{\cellcolor{LightBlue}{\textbf{ 1. Proprietary Models}}} \\
\midrule
GPT-4o~\cite{gpt4_report}$^\ddagger$ & 5 & $\sim$450 & 85 & 95.5 & 35.4 & 0.295 & 5.6 & 16.50 & \texttt{N/A} \\
Claude 3.7 Sonnet~\cite{claude3.7}$^\ddagger$ & 5 & $\sim$420 & 80 & 94.8 & 38.2 & 0.288 & 5.5 & 15.80 & \texttt{N/A} \\
Gemini 2.5 Pro~\cite{gemini2.5pro}$^\ddagger$ & 16 & $\sim$400 & 75 & 94.5 & 40.6 & 0.281 & 5.4 & 15.65 & \textbf{100} \\
o4-mini~\cite{o4mini}$^\ddagger$ & 13 & $\sim$350 & 65 & 93.0 & 45.1 & 0.270 & 5.2 & 14.50 & \textbf{100} \\
\midrule
\multicolumn{10}{l}{\cellcolor{LightBlue}{\textbf{ 2. Open-Source LLMs}}} \\
\midrule
DeepSeek-R1~\cite{deepseekr1_guo_2025} & 21 & $\sim$380 & 90 & 92.5 & 32.5 & 0.290 & 5.3 & 16.20 & \textbf{100} \\
Qwen2.5-VL-72B-Instruct~\cite{qwen2.5vl_bai_2025} & 4& $\sim$400 & 95 & 92.8 & 34.3 & 0.292 & 5.4 & 16.30 & 91.8 \\
\midrule
\multicolumn{10}{l}{\cellcolor{LightBlue}{\textbf{ 3. Optimization-based Methods}}} \\
\midrule
VectorFusion~\cite{vectorfusion_jain_2023} & 680 & 100k & 2500 & \textbf{100} & 25.0 & 0.301 & 5.7 & 18.00 & \texttt{N/A} \\
DiffSketcher~\cite{diffsketcher_xing_2023} & 550 & 100k & 2500 & \textbf{100} & 28.3 & 0.305 & 5.6 & 17.80 & \texttt{N/A} \\
SVGDreamer~\cite{svgdreamer_xing_2023} & 1020 & 100k & 2500 & \textbf{100} & 22.5 & 0.309 & 5.8 & 18.50 & \texttt{N/A} \\
\midrule
\multicolumn{10}{l}{\cellcolor{LightBlue}{\textbf{ 4. LLM-based Methods}}} \\
\midrule
LLM4SVG (Qwen2.5-7B-Instruct)~\cite{llm4svg_xing_2024} & 25 & $\sim$215 & 45 & 76.0 & 30.7 & 0.293 & 5.2 & 16.80 & \texttt{N/A} \\
StarVector (SD sampling + Img2SVG)~\cite{starvector_Rodriguez_2023} & 90 & $\sim$370 & 70 & 72.0 & 35.8 & 0.288 & 5.1 & 16.00 & \texttt{N/A} \\
\midrule
\multicolumn{10}{l}{\cellcolor{LightBlue}{\textbf{ Our Methods}}} \\
\midrule
SFT-vanilla & \textbf{5} & $\sim$300 & 55 & 75.0 & 28.1 & 0.285 & 5.3 & 17.50 & \texttt{N/A} \\
SFT-DwT (w/o RL) & 8 & $\sim$1500 & 120 & 89.0 & 21.2 & 0.310 & 5.7 & 19.50 & 92.3 \\
Reason-SVG (Full) & 12 & $\sim$3200 & 145 & 99.8 & \textbf{18.6} & \textbf{0.345} & \textbf{5.9} & \textbf{21.80} & \textbf{100} \\
\bottomrule
\end{tabular}%
}
\vspace{-0.5em}
\caption{
\textbf{Quantitative Evaluations.}  
``$\uparrow$'' and ``$\downarrow$'' indicate that higher or lower values are better, respectively.  
Evaluation metrics include Fréchet Inception Distance (FID), CLIP Text-Image Score (CLIPScore), Improved Aesthetic Predictor Score (Aesthetic), Human Preference Score (HPSv2), SVG Validity (Val\%), and DwT Adherence (DwT-Cover\%).  
``Time (s)'' denotes the average time required to generate one SVG, measured in seconds.  
``\#Token'' refers to the length of the generated SVG code after tokenization by the Qwen2.5 tokenizer.  
``\#Complex'' represents the average number of path commands and primitives in the generated SVG code.  
$^\ddagger$~Indicates results obtained via API in a zero-shot setting.  
\texttt{N/A}: Not applicable.
}
\label{tab:quantitative_evaluations}
\end{table*}
\section{Experiments}
\label{sec:experiments}

\subsection{Experimental Setup}
\label{sec:exp_setup}

\noindent\textbf{Datasets.}
We train on three datasets spanning supervised fine-tuning (SFT), reinforcement learning (RL), and evaluation. We use SVGX-SFT~\cite{llm4svg_xing_2024} for initial grounding and our SVGX-DwT-10k for Drawing-with-Thought (DwT) supervision (details in Appendix~\ref{suppsec:exp_setup_details} \&~\ref{suppsec:svgx_dwt_dataset}). A subset of $2{,}000$ prompts is used for RL ($\mathcal{D}_{\text{RL-Prompt}}$), and a held-out $1{,}000$-prompt set forms the evaluation benchmark $\mathcal{D}_{\text{Eval}}$, with no cross-phase overlap.

\noindent\textbf{Baselines.}
We compare \textbf{Reason-SVG} against: (i) general-purpose multimodal LLMs (e.g., GPT-4o, Claude 3.7 Sonnet, Gemini 2.5 Pro), (ii) open-source VLMs (e.g., DeepSeek-R1, Qwen2.5-VL-72B-Instruct), (iii) optimization-based vector graphics methods (VectorFusion, DiffSketcher, SVGDreamer), and (iv) LLM-based SVG generators (StarVector, LLM4SVG). Full model list and settings are in Appendix~\ref{suppsec:exp_setup_details}.

\noindent\textbf{Implementation.}
All experiments use Qwen2.5VL-7B-Instruct~\cite{qwen2.5vl_bai_2025} as the base model. We perform SFT on SVGX-SFT and SVGX-DwT-10k, then apply GRPO-based RL on $\mathcal{D}_{\text{RL-Prompt}}$ with a hybrid reward combining text relevance, rendering quality, semantic alignment, and aesthetics. Complete hyperparameters and training details are in Appendix~\ref{suppsec:exp_setup_details}.

\noindent\textbf{Evaluation Metrics.}
We report automatic and human evaluations. Automatic metrics include SVG Validity, CLIP-based Semantic Alignment, Aesthetic Quality, Visual Realism (FID), and DwT Adherence for models producing reasoning. Formal definitions and computation protocols are provided in Appendix~\ref{suppsec:exp_setup_details}.

\subsection{Quantitative Results and Analysis}
\label{sec:main_results}
Table~\ref{tab:quantitative_evaluations} reports the performance of \bolditn{Reason-SVG} and all baselines across six automatic evaluation metrics.  
Reason-SVG consistently outperforms both general-purpose LLMs and specialized SVG generators in semantic alignment, structural validity, visual realism, and aesthetic quality.
Among \textit{proprietary models} (GPT-4o~\cite{gpt4_report}, Claude 3.7~\cite{claude3.7}, Gemini 2.5 Pro~\cite{gemini2.5pro}, o4-mini~\cite{o4mini}), average CLIPScore remains modest ($0.289$), and visual realism lags behind (FID: $37.33$ avg.). While their SVGs are mostly valid (Val\%: $94.5$ avg.), they lack structural reasoning and exhibit no DwT adherence. These results reflect strong zero-shot generation capacity but limited control and consistency.
\textit{Open-source LLMs} (DeepSeek-R1~\cite{deepseekr1_guo_2025}, Qwen2.5-VL~\cite{qwen2.5vl_bai_2025}) slightly outperform proprietary models in both FID ($33.4$ avg.) and CLIPScore ($0.291$ avg.). More importantly, they show strong structural compliance, with an average DwT Coverage of $95.9\%$, demonstrating their capacity for controllable generation when prompted appropriately.
\textit{Optimization-based methods} (VectorFusion~\cite{vectorfusion_jain_2023}, DiffSketcher~\cite{diffsketcher_xing_2023}, SVGDreamer~\cite{svgdreamer_xing_2023}) achieve the best visual realism (FID: $25.3$ avg.) and high CLIPScore ($0.305$ avg.), confirming their strength in low-level visual quality. However, their inference time exceeds $750$ seconds on average, and token length exceeds $100k$, making them impractical for real-time or scalable applications.
\textit{LLM-based methods} (LLM4SVG~\cite{llm4svg_xing_2024}, StarVector~\cite{starvector_Rodriguez_2023}) offer lightweight inference and fast generation, but suffer from weak FID ($33.3$ avg.), lower SVG validity (Val\%: $74\%$ avg.), and lack of intermediate reasoning. They highlight the trade-off between efficiency and structural control.
By contrast, \textit{Reason-SVG} achieves the best performance across all key metrics: lowest FID ($18.6$), highest CLIPScore ($0.345$), highest HPSv2 ($21.80$), best Aesthetic score ($5.9$), near-perfect SVG Validity ($99.8\%$), and full DwT Coverage ($100\%$). Despite its multi-stage reasoning, it maintains interactive generation time ($12$ s). These results demonstrate that Reason-SVG offers the best trade-off across realism, semantic fidelity, structural correctness, and reasoning integration.

\begin{figure*}[t]
\centering
\includegraphics[width=1.0\textwidth]{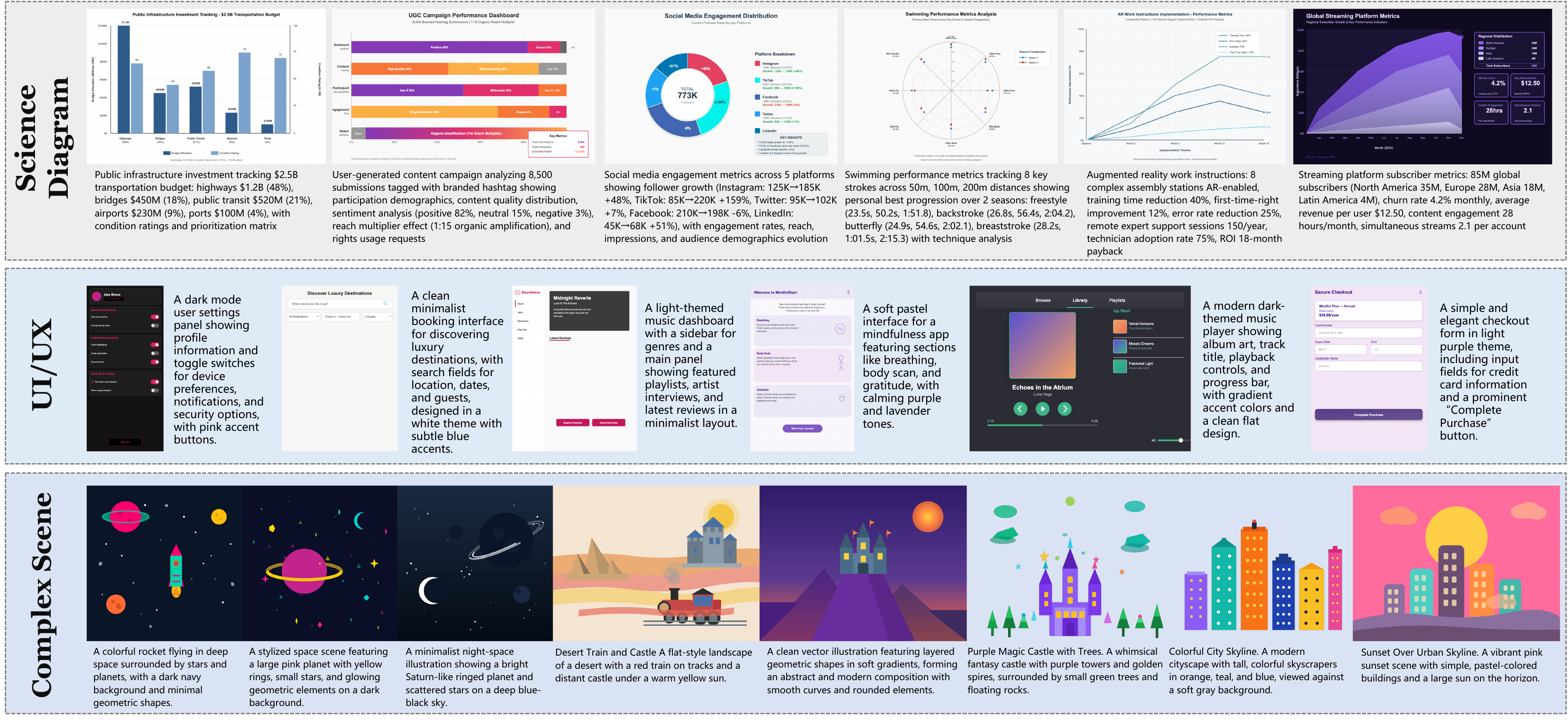}
\vspace{-1.5em}
\caption{
\textbf{Qualitative results of Reason-SVG.}
For science diagrams, the model follows the instruction “drawing an SVG-format diagram following {prompt}” to generate structured plots and analytic charts. Across diverse SVG categories—including Science Diagram, UI/UX, and Complex Scene—Reason-SVG exhibits strong visual reasoning and structural understanding. The proposed DwT reasoning further enables more coherent layout planning and significantly improves generation quality, especially in complex scenes.
The model accurately places text, axes, and semantic groups in data-driven diagrams, and preserves UI hierarchy such as panels, controls, and typography. In complex scenes, Reason-SVG organizes foreground and background elements with consistent geometry and color composition, producing clean, editable vector graphics.
}
\label{fig:complex_svg_type}
\vspace{-1em}
\end{figure*}
\begin{table}[t]
\centering
\resizebox{1.0\linewidth}{!}{%
\begin{tabular}{lccc}
\toprule
Method & SemAcc $\uparrow$ & VisApp $\uparrow$ & DwT-Qual $\uparrow$ \\
\midrule
VectorFusion~\cite{vectorfusion_jain_2023} & $3.65 \pm 0.47$ & $3.85 \pm 0.49$ & \texttt{N/A} \\
SVGDreamer~\cite{svgdreamer_xing_2023} & $3.60 \pm 0.45$ & $3.81 \pm 0.48$ & \texttt{N/A} \\
SFT-vanilla & $3.21 \pm 0.55$ & $3.05 \pm 0.60$ & \texttt{N/A} \\
LLM4SVG~\cite{llm4svg_xing_2024} & $3.48 \pm 0.49$ & $3.32 \pm 0.53$ & \texttt{N/A} \\
GPT-4o~\cite{gpt4_report}$^\ddagger$ & $3.75 \pm 0.48$ & $3.60 \pm 0.52$ & \texttt{N/A} \\
\midrule
SFT-DwT (w/o RL) & $3.95 \pm 0.42$ & $3.70 \pm 0.51$ & $3.92 \pm 0.38$ \\
\bolditn{Reason-SVG} (Full) & $\mathbf{4.53 \pm 0.35}$ & $\mathbf{4.42 \pm 0.39}$ & $\mathbf{4.61 \pm 0.31}$ \\
\bottomrule
\end{tabular}
}
\vspace{-0.5em}
\caption{
Human Evaluation Results (mean scores $\pm$ std. dev. on 1--5 Likert scale). 
}
\label{tab:human_eval}
\vspace{-2.0em}
\end{table}

\noindent\textbf{Human Evaluation.}
We conducted a human evaluation study to assess the quality of generated SVGs and their underlying reasoning.  
A total of 19 participants with backgrounds in graphic design and visual communication rated model outputs based on a randomly sampled subset of 50 prompts from the held-out evaluation set $\mathcal{D}_{\text{Eval}}$.
Participants scored each result along three dimensions using a 1–5 Likert scale:  
(1) \textit{Semantic Accuracy} (SemAcc), measuring how accurately the SVG reflects the intended meaning of the prompt;  
(2) \textit{Visual Appeal} (VisApp), evaluating the perceived aesthetic quality of the SVG; and  
(3) \textit{DwT Quality} (DwT-Qual), applicable to models producing intermediate reasoning, which assesses the coherence, logical structure, and task relevance of the generated Drawing-with-Thought sequence.  
All model sources were anonymized and their presentation order randomized. 

The results from our human evaluation study are presented in Table~\ref{tab:human_eval}. Reason-SVG significantly outperforms all baselines in Semantic Accuracy and Visual Appeal.  
Importantly, its generated DwT sequences also receive high ratings for quality ($4.61 \pm 0.31$), validating the utility of explicit intermediate reasoning.  
In pairwise comparisons against the strongest baseline (SVGDreamer~\cite{svgdreamer_xing_2023}), outputs from Reason-SVG were preferred 78\% of the time.  
These results highlight Reason-SVG’s ability to produce SVGs that better align with human perception and design intent.

\begin{figure*}[t]
\centering
\includegraphics[width=1.0\textwidth]{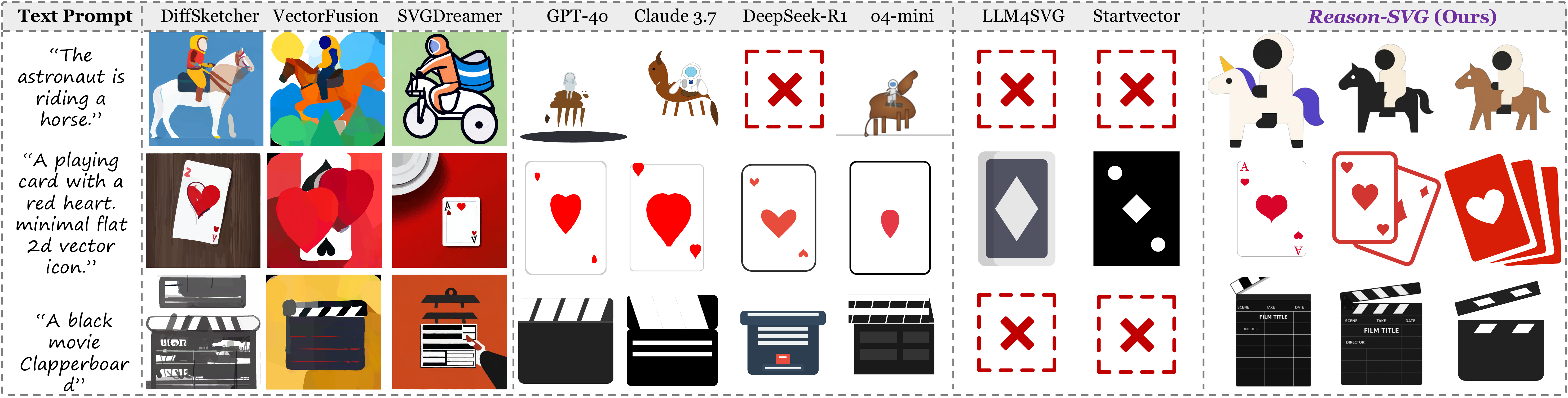}
\vspace{-1.5em}
\caption{
\textbf{Qualitative results on diverse prompts.} 
We evaluate Reason-SVG against both optimization-based methods (DiffSketcher, VectorFusion, SVGDreamer) and LLM-based baselines (GPT-4o, Claude 3.7, DeepSeek-R1, o4-mini, LLM4SVG, StarVector) on a diverse set of prompts spanning scientific objects, cultural icons, animals, compositional scenes, and abstract symbols. Reason-SVG consistently generates clean, well-structured, and semantically accurate vector graphics that better match the prompt intent, especially in cases requiring compositional reasoning or precise symbolic representation.
}
\label{fig:qualitative}
\vspace{-0.5em}
\end{figure*}

\subsection{Qualitative Results and Analysis}
Figure~\ref{fig:complex_svg_type} showcases \textbf{Reason-SVG} results across Science Diagram, UI/UX, and Complex Scene categories. The outputs exhibit clean geometry, coherent layout, and consistent styling—evidence that DwT planning improves structure in non-icon settings.

Figure~\ref{fig:qualitative} provides side-by-side comparisons with optimization-based~\cite{diffsketcher_xing_2023,vectorfusion_jain_2023,svgdreamer_xing_2023}, proprietary~\cite{gpt4_report,claude3.7}, open-source~\cite{deepseekr1_guo_2025,qwen2.5vl_bai_2025}, and LLM-based SVG generators~\cite{llm4svg_xing_2024,starvector_Rodriguez_2023}. 
On single-object prompts (e.g., ``an icon of the planet Saturn'', ``the Statue of Liberty''), Reason-SVG preserves distinctive shapes and proportions, while baselines tend to oversimplify or distort. 
For compositional prompts (e.g., ``the astronaut is riding a horse''), Reason-SVG correctly integrates entities with proper spatial relations and recognizable silhouettes; many baselines miss parts or misalign objects. 
For symbolic designs (e.g., ``a playing card with a red heart'', ``a black movie clapperboard''), our outputs follow expected visual conventions with valid SVG structure.

Overall, the DwT pipeline yields stronger compositional reasoning and layout fidelity, complementing the broader, more complex categories demonstrated in Figure~\ref{fig:complex_svg_type}.

\begin{table}
\centering
\resizebox{1.0\linewidth}{!}{
\begin{tabular}{lcccc}
\toprule
Variant & CLIPScore $\uparrow$ & HPSv2 $\uparrow$ & Val \% $\uparrow$ & DwT-Cover \% $\uparrow$ \\
\midrule
Full Reason-SVG (Ablation Baseline) & \textbf{0.345} & \textbf{21.40} & \textbf{97.8} & \textbf{100} \\
\midrule
\multicolumn{5}{l}{ \cellcolor{LightBlue}{\textbf{Impact of DwT}} } \\
Reason-SVG w/o DwT (\& w/o $\mathcal{R}_{\text{think}}$) & 0.304 & 18.42 & \texttt{N/A} & \texttt{N/A} \\
\midrule
\multicolumn{5}{l}{ \cellcolor{LightBlue}{\textbf{Impact of Hybrid Reward (all include DwT-SFT \& RL)}} } \\
w/o $\mathcal{R}_{\text{think}}$ & 0.313 & 20.15 & 97.1 & 85.3 \\
w/o $\mathcal{R}_{\text{render}}$ & 0.328 & 20.95 & 82.5 & 95.8 \\
w/o $\mathcal{R}_{\text{semantic}}$ & 0.289 & 20.50 & 97.5 & 98.1 \\
w/o $\mathcal{R}_{\text{aesthetic}}$ & 0.341 & 18.25 & 97.6 & 100 \\
\bottomrule
\end{tabular}%
}
\vspace{-0.6em}
\caption{
Ablation studies on the impact of ``Drawing-with-Thought'' (DwT) and hybrid reward components. Baseline ``Full Reason-SVG'' values are specific to this ablation setup.
}
\label{tab:ablation_combined} 
\vspace{-1.0em}
\end{table}
\subsection{Ablation Studies}
\label{sec:ablation_studies}
\textbf{Impact of} \bolditn{Drawing-with-Thought}.
We compare our proposed Reason-SVG with a variant where the SFT phase does not involve DwT, \textit{i.e.}, SFT-vanilla, and the RL stage excludes the $\mathcal{R}_{\text{think}}$ component. 
As shown in Table~\ref{tab:ablation_combined}, removing DwT leads to a significant drop in performance: CLIPScore decreases from $0.345$ to $0.304$, and HPSv2 drops from $21.40$ to $18.42$. These results indicate that incorporating the DwT stage—explicitly encouraging reasoning before drawing—is crucial for producing semantically meaningful and aesthetically superior SVGs. The substantial gains observed validate the importance of thoughtful reasoning in the generation process.

\noindent\textbf{Efficacy of the Reinforcement Learning Stage.}
By comparing SFT-DwT (Ours, w/o RL) with the full Reason-SVG (see Table~\ref{tab:quantitative_evaluations}), we observe that the RL stage with our hybrid reward yields an improvement from $0.310$ to $0.345$ in CLIPScore and from $19.50$ to $21.80$ in HPSv2. This demonstrates the effectiveness of RL in refining the policy learned during SFT.

\noindent\textbf{Contribution of Hybrid Reward Function.}
We analyze the impact of each component in our hybrid reward function (Eq.~\ref{eq:hyper_reward}) by training Reason-SVG variants where one reward term (and its weight) is removed at a time.  
As shown in Table~\ref{tab:ablation_combined}, removing any component leads to a noticeable degradation in performance compared to the `Full Reason-SVG (Ablation Baseline)'' row. For instance, without $\mathcal{R}_{\text{think}}$, DwT-Cover drops by $11.2$ percentage points (from $96.5\%$ to $85.3\%$) and CLIPScore by $0.032$, highlighting the importance of explicitly rewarding coherent thought processes.  
Similarly, removing $\mathcal{R}_{\text{aesthetic}}$ results in a lower HPSv2 score by $3.15$ (from $21.40$ to $18.25$).

These results validate the effectiveness of our reward design.  
Notably, the performance drop varies across metrics, indicating that each component targets a distinct yet complementary aspect of generation quality.  
The hybrid reward plays a critical role in balancing low-level structure with high-level semantics and perceptual appeal—essential for reasoning-driven SVG synthesis.

\section{Conclusion \& Discussion}


We present Reason-SVG, a framework that advances LLM-based SVG generation through reasoning-driven synthesis, where explicit visual planning guides the creation of structured vector graphics. At the core is Drawing-with-Thought (DwT), which prompts the model to articulate semantic, structural, and aesthetic decisions before producing SVG code.

By integrating DwT-supervised fine-tuning with reinforcement learning using a Hybrid Reward, Reason-SVG achieves substantial improvements in semantic alignment, structural validity, and visual quality. These results show that explicit reasoning offers a powerful intermediate representation for vector graphics generation. This paradigm also opens pathways toward broader reasoning-guided multimodal creation, including more complex vector formats, interactive editing, and tighter coupling with perceptual feedback.

Beyond performance gains, our findings highlight a more general insight: introducing structured reasoning can fundamentally reshape how LLMs interpret, plan, and execute visual design tasks. We believe this direction can benefit downstream applications such as diagram synthesis, UI layout planning, and vector editing agents, and may inspire future research into models that unify symbolic structure with continuous visual understanding.

\clearpage
\renewcommand{\thefigure}{S\arabic{figure}}
\setcounter{figure}{0}
\renewcommand{\thetable}{S\arabic{table}}
\setcounter{table}{0}
\maketitlesupplementary

\appendix

\section*{Overview}
\label{sec:overview}

This supplementary material provides additional implementation details, dataset statistics, and qualitative analyses to support the findings of the main paper. The content is organized as follows:

\begin{itemize}
\setlength{\itemsep}{0.5em} 
\item \cref{suppsec:exp_setup_details}: \textbf{Experimental Setup: Full Details.} Comprehensive protocols for training and evaluation, including detailed baseline configurations, extended implementation specifics, and definitions of automatic metrics.
\item \cref{suppsec:svgx_dwt_dataset}: \textbf{SVGX-DwT-10k Dataset.} An in-depth look at the \textit{automated, VLM-verified} construction pipeline, statistical analysis of reasoning depth, and diverse examples across domains.
\item \cref{suppsec:dwt_details}: \textbf{DwT Case Study.} A step-by-step visualization of the Drawing-with-Thought reasoning process, illustrating the progression from concept sketching to final execution.
\item \cref{suppsec:img2svg}: \textbf{Extension to Image-to-SVG Generation.} Qualitative and quantitative evaluation of Reason-SVG's capability in vectorization tasks, demonstrating how the multimodal backbone facilitates structural reconstruction from raster inputs.
\end{itemize}

\begin{figure*}[t]
\centering
\includegraphics[width=1.0\textwidth]{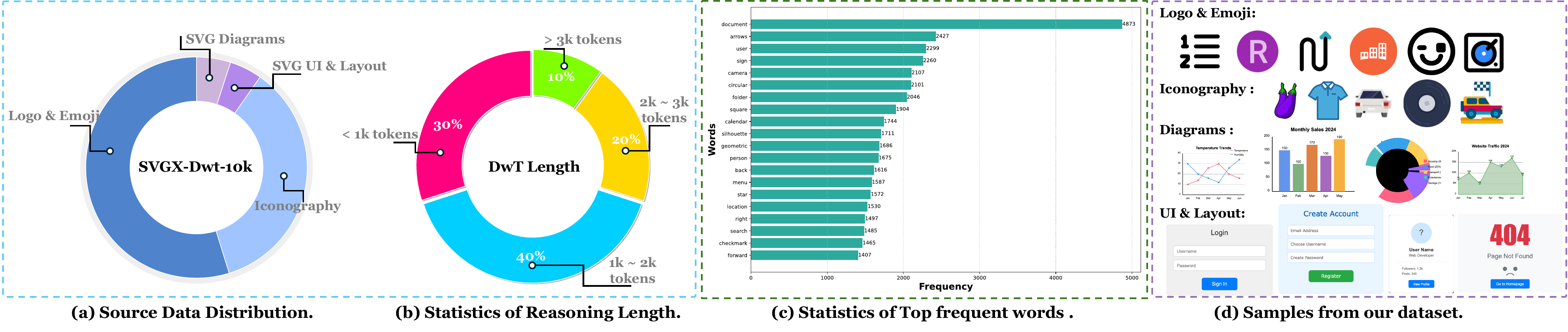}
\vspace{-1.5em}
\caption{
\textbf{Statistical Overview of the SVGX-DwT-10k Dataset.} 
\textbf{(a) Domain Distribution:} The dataset comprises four distinct categories, with a predominance of icon-centric graphics (Logo \& Emoji, Iconography) complemented by structured Diagrams and UI layouts.
\textbf{(b) Reasoning Depth:} Distribution of DwT sequence lengths. Notably, 70\% of samples exceed 1k tokens, indicating that the dataset captures comprehensive, multi-step planning rather than superficial descriptions.
\textbf{(c) Semantic Vocabulary:} Frequency analysis of the top concepts in the reasoning traces, highlighting common structural primitives (e.g., \textit{document}, \textit{user}, \textit{geometric}).
\textbf{(d) Qualitative Samples:} Representative triplets showcasing the diversity in style, complexity, and structural logic across different domains.
}
\label{fig:svgx_dwt_meta_info}
\end{figure*}
\section{Experimental Setup: Full Details}
\label{suppsec:exp_setup_details}

In this section, we provide the comprehensive protocols used for benchmarking and the granular implementation details of our proposed framework to facilitate reproducibility.

\subsection{Baseline Methods}
We benchmark \bolditn{Reason-SVG} against a diverse set of baselines spanning general-purpose LLMs, open-source models, optimization-based techniques, and LLM-based SVG generators. Specifically, we evaluate:
(1) \textit{General-purpose LLMs}, including GPT-4o~\cite{gpt4_report}, Claude 3.7 Sonnet~\cite{claude3.7}, and Gemini 2.5 Pro~\cite{gemini2.5pro};
(2) \textit{Open-source Models}, specifically DeepSeek-R1~\cite{deepseekr1_guo_2025} and Qwen2.5-VL-72B-Instruct~\cite{qwen2.5vl_bai_2025};
(3) \textit{Optimization-based Methods}, namely VectorFusion~\cite{vectorfusion_jain_2023}, DiffSketcher~\cite{diffsketcher_xing_2023}, and SVGDreamer~\cite{svgdreamer_xing_2023}; and
(4) \textit{LLM-based SVG Generators}, including StarVector~\cite{starvector_Rodriguez_2023} and LLM4SVG~\cite{llm4svg_xing_2024}.

\subsection{Extended Implementation Details}
All experiments utilize Qwen2.5-VL-7B-Instruct~\cite{qwen2.5vl_bai_2025} as the foundational model. We selected this Vision-Language Model (VLM) for its state-of-the-art instruction-following capabilities and its native visual encoder, which facilitates the seamless extension to Image-to-SVG tasks without requiring additional adapters. For Text-to-SVG generation, the visual input channel is masked, allowing the model to function effectively as a text-only generator.

The \textbf{Supervised Fine-Tuning (SFT)} phase utilized both the existing SVGX-SFT dataset~\cite{llm4svg_xing_2024} and our novel SVGX-DwT-10k dataset. SFT was conducted for 3 epochs with a global batch size of 32. The AdamW optimizer was employed with standard beta values ($\beta_1 = 0.9, \beta_2 = 0.999$) and a weight decay $\epsilon = 10^{-8}$. A peak learning rate of $2 \times 10^{-5}$ was used, coupled with a cosine decay schedule and a warm-up phase constituting approximately 10\% of the initial training steps. Input sequences were tokenized and truncated or padded to a maximum sequence length of 4096 tokens, ensuring compatibility with the model's context window.

Following SFT, the \textbf{Reinforcement Learning (RL)} phase leveraged Group Relative Policy Optimization (GRPO)~\cite{deepseekr1_guo_2025} to refine the model. Unlike standard PPO, GRPO estimates advantages by comparing multiple trajectories sampled from the current policy rather than relying on a learned value function, making it highly stable for our rule-based reward setting. Training was performed for 8000 policy update steps using the $\mathcal{D}_{\text{RL-Prompt}}$ dataset. Key GRPO hyperparameters included a group size $G = 8$ for trajectory sampling. The advantage $\hat{A}_k$ for each candidate sequence was computed per-token by comparing its total hybrid reward against the group's average performance. A PPO-style clipping parameter $\epsilon = 0.2$ was used in the surrogate objective, and a Kullback-Leibler (KL) divergence penalty with a coefficient $\beta = 0.01$ was applied to regularize the policy updates against a reference policy. The reference policy was updated via an exponential moving average with a decay rate of 0.99. Individual reward components from the hybrid reward function were weighted by coefficients $\lambda_t = 0.1$ (Text Relevance), $\lambda_r = 0.1$ (Rendering Quality), $\lambda_s = 0.6$ (Semantic Alignment), and $\lambda_a = 0.2$ (Aesthetics), and normalized to a consistent range before summation to ensure training stability.

All experiments were executed on a cluster of 32 NVIDIA H800 (80GB) GPUs, with distributed training managed using standard data parallelism. 
CairoSVG~\cite{cairosvg} was used for SVG rendering and validation. Semantic alignment scoring utilized the official OpenAI CLIP library (ViT-L/14 model)~\cite{clip_Radford_2021}, while aesthetic assessment employed the HPSv2~\cite{HPS_Wu_2023} model implementation.

\subsection{Evaluation Metrics}
We employ both automatic and human evaluations to assess the quality of generated SVGs.

\noindent\textbf{SVG Validity (Val\%)} measures the proportion of outputs that are syntactically correct and successfully rendered using CairoSVG~\cite{cairosvg}.

\noindent\textbf{Semantic Alignment} is evaluated using CLIPScore, computed as the cosine similarity between CLIP ViT-L/14~\cite{clip_Radford_2021} embeddings of the rendered SVG image and the input prompt. Formally, $\text{CLIPScore}=\cos\!\big(\phi_{\text{img}}(I),\,\phi_{\text{text}}(T)\big).$

\noindent\textbf{Aesthetic Quality} is assessed via HPSv2~\cite{HPS_Wu_2023}, which predicts a score of human-perceived visual appeal based on the rendered image.

\noindent\textbf{Visual Realism} is measured using Fréchet Inception Distance (FID) between the distribution of rendered SVGs and natural icon distributions, where lower values indicate better realism.

\noindent\textbf{DwT Adherence (DwT-Cover\%)} calculates the proportion of outputs that contain a structurally valid DwT sequence (i.e., correct tags and stages), applicable only to models producing intermediate reasoning.

\noindent\textbf{Structural Complexity (\#Complex)} represents the average number of path commands and primitive elements in the generated SVG code. Concretely, for each SVG $i$, we count: (1) Path commands in all \texttt{<path>} elements (e.g., M, L, H, V, C, S, Q, T, A, Z); and (2) Primitive elements (e.g., \texttt{<rect>}, \texttt{<circle>}, \texttt{<ellipse>}, \texttt{<line>}, \texttt{<polyline>}, \texttt{<polygon>}). We then average this count over the evaluation set:
\begin{equation}
\#\text{Complex} \;=\; \frac{1}{N}\sum_{i=1}^{N} \big(|\mathcal{C}_i| + |\mathcal{P}_i|\big),
\end{equation}
where $|\mathcal{C}_i|$ is the total number of parsed path commands and $|\mathcal{P}_i|$ is the number of primitive elements in SVG $i$. Higher values indicate more structurally complex SVGs; this metric is reporting-only and does not affect training.

\begin{figure*}[t]
\centering
\includegraphics[width=1.0\textwidth]{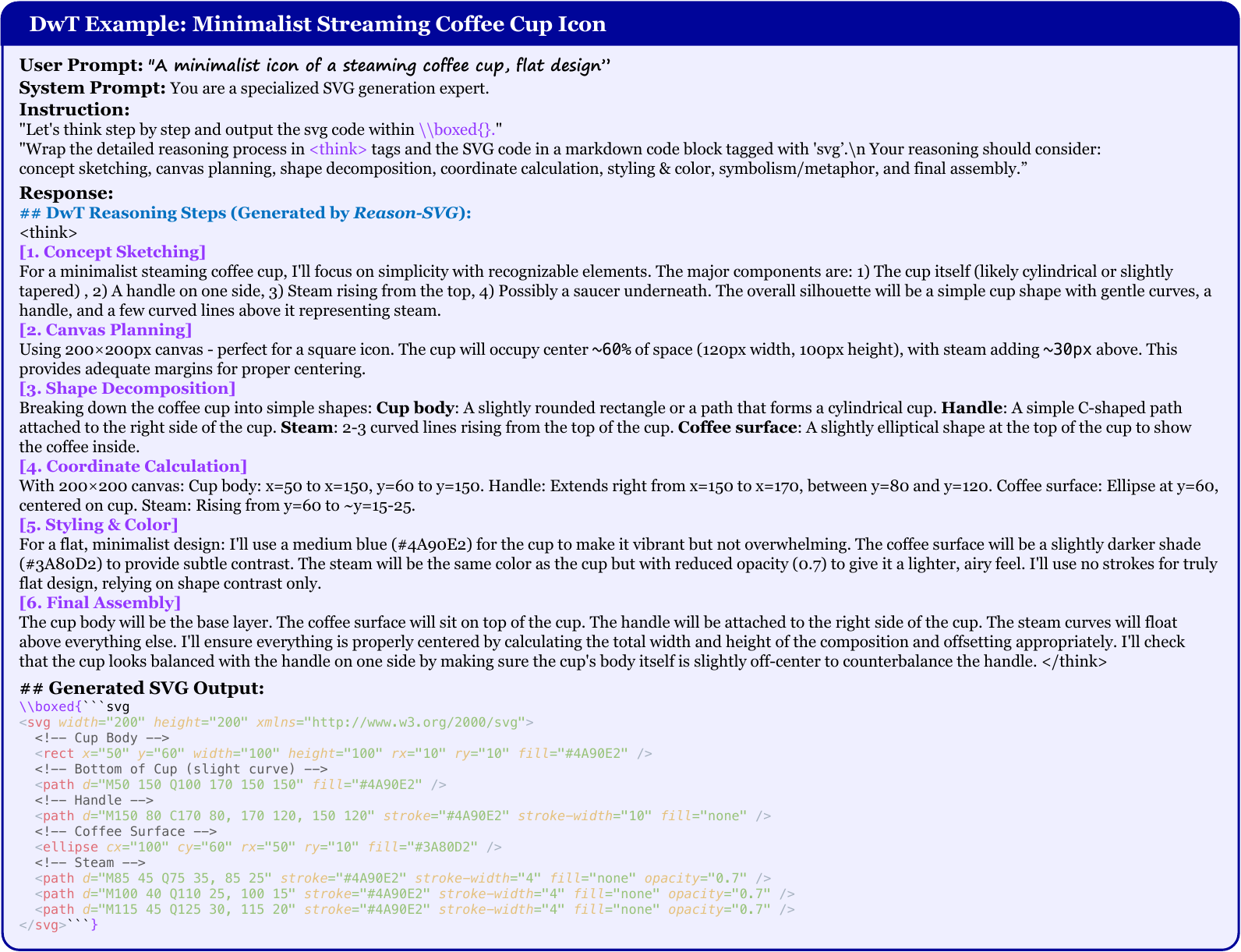}
\vspace{-1em}
\caption{
\textbf{DwT Example: Minimalist Streaming Coffee Cup Icon.}  
This figure showcases the full pipeline for generating an SVG via the ``Drawing-with-Thought'' (DwT) paradigm.  
It includes the user prompt, system instruction, and the six-stage reasoning process generated by \bolditn{Reason-SVG}, covering concept sketching, canvas planning, shape decomposition, coordinate calculation, styling \& color, and final assembly.  
The corresponding SVG code and rendered visual output are also shown, offering a complete illustration of structured SVG generation.
}
\label{fig:dwt_example}
\vspace{-1.5em}
\end{figure*}

\section{SVGX-DwT-10k Dataset}
\label{suppsec:svgx_dwt_dataset}

The SVGX-DwT-10k dataset constitutes the foundational asset of our framework, bridging the gap between abstract natural language prompts and executable vector code through explicit intermediate reasoning. As illustrated in \cref{fig:svgx_dwt_meta_info}, the dataset comprises $10,000$ high-quality triplets $(\mathcal{T}, C, O)$, where $\mathcal{T}$ denotes the textual prompt, $C$ represents the structured Drawing-with-Thought (DwT) rationale, and $O$ is the resulting SVG code. 

\noindent\textbf{Domain Diversity and Composition.} 
To ensure robust generalization across various vector graphics tasks, we curated the dataset to cover four distinct domains: \textit{Logo \& Emoji}, \textit{Iconography}, \textit{UI \& Layout}, and \textit{Diagrams} (\cref{fig:svgx_dwt_meta_info}(a)). While icon-style graphics (Logo/Iconography) form the majority ($>80\%$) to support object-centric reasoning, the inclusion of UI components and analytical charts introduces critical challenges related to layout constraints, text rendering, and hierarchical grouping. This structural diversity prevents the model from overfitting to simple single-object generation.

\noindent\textbf{Reasoning Depth and Vocabulary.} 
A key differentiator of SVGX-DwT-10k is the depth of its reasoning traces. As shown in \cref{fig:svgx_dwt_meta_info}(b), $70\%$ of the DwT sequences exceed $1,000$ tokens, with $10\%$ surpassing $3,000$ tokens. This length distribution reflects the rigorous six-stage planning process—spanning concept sketching to coordinate calculation—required to generate valid SVGs. Furthermore, the vocabulary analysis in \cref{fig:svgx_dwt_meta_info}(c) reveals that the reasoning process is grounded in specific semantic primitives (e.g., \textit{document}, \textit{arrow}, \textit{circular}, \textit{geometric}), confirming that the DwT traces focus on actionable design elements rather than generic conversational filler.

\noindent\textbf{Automated Construction Pipeline.} 
To ensure the alignment between the reasoning trace $C$ and the code $O$ at scale, we established a rigorous \textit{Generate-Render-Verify} pipeline powered by the multimodal capabilities of Gemini 2.5 Pro:
\begin{enumerate}
    \item \textit{Structured Generation:} We prompted Gemini 2.5 Pro~\cite{gemini2.5pro} with a specialized system instruction to generate the triplet $(\mathcal{T}, C, O)$, strictly enforcing the six-stage DwT format.
    \item \textit{Execution \& Syntax Validation:} The generated SVG code $O$ was compiled using CairoSVG~\cite{cairosvg}. Samples causing rendering errors, parsing failures, or resulting in empty canvasses were automatically discarded.
    \item \textit{VLM-based Consistency Filtering:} A critical challenge in synthetic data is hallucination, where the code ignores the reasoning plan. To address this, we implemented a \textit{Visual-Reasoning Consistency Check}. 
    We fed the rendered raster image $I(O)$, the original prompt $\mathcal{T}$, and the reasoning trace $C$ back into Gemini 2.5 Pro. Acting as a critic, the model evaluated whether the visual output faithfully realized the design decisions described in $C$ (e.g., ``Does the image actually contain the red circle mentioned in the Shape Decomposition step?''). Only samples rated as highly consistent were retained for the final corpus.
\end{enumerate}

\begin{figure*}[h]
\centering
\includegraphics[width=1.0\textwidth]{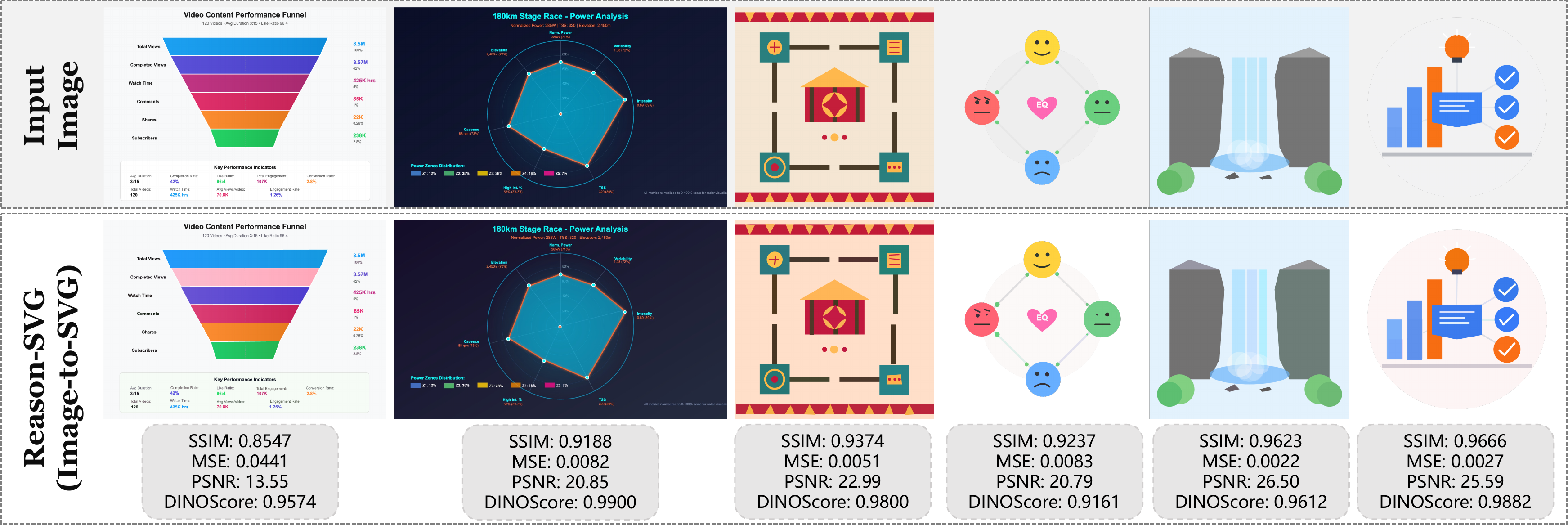}
\vspace{-1.5em}
\caption{
\textbf{Qualitative and quantitative results of Reason-SVG on the Image-to-SVG task.}  
The top row displays the input raster images, while the bottom row presents the rendered SVG outputs generated by our method. The examples demonstrate the model's versatility across diverse domains, including data visualization (funnel and radar charts), structured layouts, and flat illustrations. Quantitative metrics (SSIM, MSE, PSNR, and DINOScore) are reported below each column. The high consistency between the inputs and outputs, evidenced by high SSIM and DINOScore values, highlights Reason-SVG's capability to achieve high-fidelity vectorization with precise structural and semantic preservation.
}
\label{fig:img2svg}
\end{figure*}

\section{Case Study: A Step-by-Step Illustration of DwT Reasoning}
\label{suppsec:dwt_details}

To further demonstrate the effectiveness of the \bolditn{Drawing-with-Thought} (DwT) mechanism introduced in Section~\ref{sec:dwt}, we present a concrete example illustrating the full SVG generation pipeline.

\cref{fig:dwt_example} details how Reason-SVG processes the prompt \textit{``A minimalist icon of a steaming coffee cup, flat design''} through six structured reasoning stages. Unlike black-box generation, our model explicitly articulates its design decisions before writing code:

\begin{itemize}
    \item \textbf{(a) Concept Sketching:} The model correctly interprets the abstract stylistic constraint ``minimalist'' by deciding to focus on a ``simple silhouette'' and explicitly listing major components (body, handle, steam).
    
    \item \textbf{(b) Canvas Planning:} It proactively defines the workspace, choosing a $200{\times}200$ canvas and allocating $\sim$60\% of the space to the cup to ensure proper margins.
    
    \item \textbf{(c) Shape Decomposition:} The complex object is broken down into geometric primitives. Notably, the model plans to represent the ``coffee surface'' as a ``slightly elliptical shape,'' anticipating the perspective needed for a 2D flat icon.
    
    \item \textbf{(d) Coordinate Calculation:} Abstract spatial relationships are grounded into concrete numbers. The model calculates specific coordinates (e.g., ``Body: x=50 to x=150''), creating a mental bounding box before any code is written.
    
    \item \textbf{(e) Styling and Color:} The model demonstrates stylistic reasoning. Recognizing the request for ``flat design,'' it explicitly decides: \textit{``No strokes used... relying on shape contrast only,''} and applies a specific opacity (0.7) to the steam elements to create a visual hierarchy.
    
    \item \textbf{(f) Final Assembly:} The layering order is logically determined (body first, then surface, then steam) to ensure correct occlusion in the rendered vector graphic.
\end{itemize}

This case study highlights the \textbf{interpretability} and \textbf{controllability} of our framework. By externalizing the visual reasoning process, Reason-SVG ensures that the final SVG code is not merely a memorized pattern, but the result of a structured, step-by-step design derivation. The high correspondence between the reasoning trace (``<think>'') and the final output validates that the model effectively adheres to its own generated plan.

\section{Extension to Image-to-SVG Generation}
\label{suppsec:img2svg}
Although \bolditn{Reason-SVG} is primarily optimized for Text-to-SVG generation, our architecture leverages the multimodal capabilities of Qwen2.5-VL-7B~\cite{qwen2.5vl_bai_2025} as a backbone. This design inherently allows the model to process raster images as input without requiring any architectural modifications. 

To activate this capability, we constructed a visual instruction tuning dataset derived from \textbf{SVGX-DwT-10k}. Specifically, we rasterized the ground-truth SVG code from the dataset into high-resolution images to form (Image, SVG) pairs. We then fine-tuned the model on these pairs using the instruction ``\textit{Recreate this image as an SVG with structured reasoning.}'' This process effectively transfers the model's reasoning capabilities from the textual to the visual domain, enabling it to analyze pixel inputs and reconstruct them as structured vector graphics.

\subsection{Quantitative Performance}
To quantitatively evaluate the reconstruction quality, we measured the agreement between the input raster images and the rendered SVG outputs across a diverse set of test cases. As illustrated in \cref{fig:img2svg}, our model achieves impressive fidelity. On average, \bolditn{Reason-SVG} attains a Structural Similarity Index (SSIM) of 0.9273 and a Peak Signal-to-Noise Ratio (PSNR) of 21.72 dB, indicating high pixel-level precision. Furthermore, the low Mean Squared Error (MSE) of 0.0118 confirms the model's ability to closely match the spatial distribution of the original image.

Beyond pixel-level metrics, we evaluated semantic preservation using the DINOScore~\cite{dinov2_oquab_2024}, which measures the cosine similarity between DINO-v2 embeddings of the input and output. The high average DINOScore of 0.9731 demonstrates that our reasoning-driven approach captures the high-level semantic identity of the visuals, ensuring that the generated SVGs are not just visually similar but semantically equivalent to the inputs.

\subsection{Role of DwT in Visual Reconstruction}
The \textit{Drawing-with-Thought} (DwT) paradigm proves pivotal in this task, distinguishing our method from traditional vectorization algorithms (e.g., Potrace) or purely end-to-end neural methods. Instead of performing blind edge-tracing, \bolditn{Reason-SVG} first ``reasons'' about the visual input:
\begin{enumerate}
\item \textbf{Visual Analysis (Concept Sketching):} The model identifies semantic components (e.g., recognizing a ``funnel chart'' or a ``radar plot'' in \cref{fig:img2svg}, rather than just seeing colored blobs).
\item \textbf{Structural Planning:} It decomposes the image into geometric primitives and infers occlusion relationships (e.g., knowing the background layer must be drawn before the foreground icons).
\item \textbf{OCR and Layout:} For data visualizations (left two columns of \cref{fig:img2svg}), the model successfully recognizes and transcribes text labels while maintaining their relative positions, a capability derived from the VLM's pre-training but structured by DwT.
\end{enumerate}

As shown in the qualitative results (\cref{fig:img2svg}), this approach allows for the reconstruction of complex diagrams, flat illustrations, and UI layouts with clean topology and editable code structure, confirming the versatility of the Reason-SVG framework.

{
    \small
    \bibliographystyle{ieeenat_fullname}
    \bibliography{main}
}


\end{document}